\documentclass{article}
\usepackage{spconf,amsmath,graphicx,url}

\title{Learning to detect keyword parts and whole by smoothed max pooling}
\name{Hyun-Jin Park, Patrick Violette, Niranjan Subrahmanya}
\address{Google Inc.\\
\texttt{\{hjpark,pdv,sniranjan\}@google.com}}
\begin{document}
\ninept
\maketitle
\begin{abstract}
We propose \emph{smoothed} max pooling loss and its application to keyword spotting systems. The proposed approach jointly trains an encoder (to detect keyword parts) and a decoder (to detect whole keyword) in a semi-supervised manner. The proposed new loss function allows training a model to detect parts and whole of a keyword, without strictly depending on frame-level labeling from LVCSR (Large vocabulary continuous speech recognition), making further optimization possible. The proposed system outperforms the baseline keyword spotting model in \cite{Alvarez2019} due to increased optimizability. Further, it can be more easily adapted for on-device learning applications due to reduced dependency on LVCSR.\end{abstract}
\section{Introduction}
\label{sec:intro}

Keyword detection has become an important frontend service for ASR-based assistant interfaces (e.g. “Hey Google”, “Alexa”, “Hey Siri”). As assistant technology spreads to more ubiquitous use-cases (mobile, IOT), reducing resource consumption (memory and computation) while improving accuracy has been the key success criteria of keyword spotting techniques. 

Following the successes in general ASR \cite{Speech12,He2018StreamingES}, the neural network based approach has been extensively explored in keyword spotting area with benefits of lowering resource requirements and improving accuracy \cite{Agc15,HeySiri17,Alexa16, Alexa17, AlexaRaw17, AlexaDelayed18, Alexa18, Hotwordv1}. Such works include DNN + temporal integration \cite{Agc15,HeySiri17,Hotwordv1,Cascade17}, and HMM + DNN hybrid approaches \cite{Alexa16, Alexa17, AlexaRaw17, AlexaDelayed18, Alexa18}. Recently introduced end-to-end trainable DNN approaches \cite{Alvarez2019,Mazzawi2019} further improved accuracy and lowered resource requirements using highly optimizable system design.

In general, training of such DNN based systems required frame-level labels generated by LVCSR systems \cite{Cnn15, Alvarez2019}. These approaches make end-to-end optimizable keyword spotting system depend on labels generated from non-end-to-end system trained for a different task. However, for keyword-spotting, the exact position of the keyword is not as relevant as its presence. Therefore, such strict dependency on frame-level labels may limit further optimization promised by the end-to-end approach.
In \cite{Alvarez2019}, the top level loss is derived by integrating frame-level losses, which are computed using frame-level labels from LVCSR. 
Integrating frame-level losses penalizes \emph{slightly mis-aligned} correct predictions, which can limit detection accuracy, especially for difficult data (e.g. \emph{noisy or accented speech}) where LVCSR labels may have higher-than-normal uncertainty.
In such case, losses can be fully minimized only when the predicted value and position-in-time matches that of provided frame level labels, where exact position match is \emph{not highly relevant} for high accuracy.


Prior work of CTC-training \cite{Ctc17} or sequence-to-sequence training \cite{Custom17} don’t require on frame level alignment information. However those approaches need to train full-sized encoders, which require fully transcribed speech data. Work in \cite{Sun2016MaxpoolingLT} proposed max pooling loss, which doesn’t depend on phoneme level alignment information, but its application is limited to decoder level training.

In this paper, we prepose a new \emph{smoothed} max pooling loss for training an end-to-end keyword spotting system. The new loss function reduces dependence on dense labels from LVCSR. Further, the new loss function jointly trains an encoder (detecting keyword parts) and decoder (detecting whole keyword). One can train models to generate stable activations for a target pattern even without exact location of the target specified. We describe the details of the proposed method in Section \ref{sec:system}. Then we show experiment setup in Section \ref{sec:setup}, and results in Section \ref{sec:results}. We conclude with discussions in Section \ref{sec:conclusion}.


\section{Smoothed max pooling loss for training encoder/decoder keyword spotting model}
\label{sec:system}

The proposed model uses the same encoder/decoder structure as \cite{Alvarez2019} (Fig.\ref{fig:topology}), but it differs in that encoder and decoder models are trained simultaneously using smoothed max pooling loss. In \cite{Alvarez2019}, both encoder and decoder models are trained with cross entropy (CE) loss using frame level labels. In the proposed approach, we define losses for encoder and decoder using smoothed max pooling loss, and optimize the combination of two losses simultaneously. The proposed smoothed max pooling loss doesn’t strictly depend on phoneme-level alignment, allowing better optimization than the baseline.

\subsection{Baseline end-to-end keyword spotting model}
\label{baseline}

\begin{figure}
	\centering
	\includegraphics[width=\columnwidth]{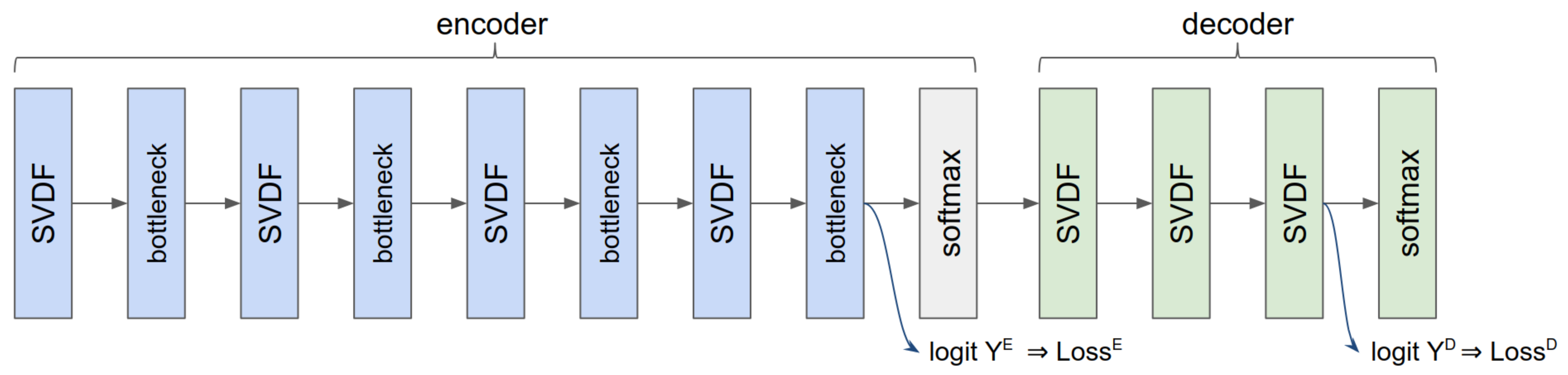}
	\caption{End-to-end topology trained to predict the keyword likelihood score. Bottleneck layers reduce parameters and computation. The intermediate softmax is used in encoder+decoder training only.}
	\label{fig:topology}
\end{figure}

Both the baseline and the proposed model have an encoder which takes spectral domain feature $\mathbf{X}_t$ as input and generate (K+1) outputs $Y^E$ corresponding to  phoneme-like sound units (Fig.\ref{fig:topology}). The decoder model takes the encoder output as input and generates binary output $Y^D$ that predicts existence of a keyword . The model is fed with acoustic input features at each frame (generated every 10ms), and generates prediction labels at each frame in a streaming manner. In \cite{Alvarez2019}, the encoder model is trained first, and then the decoder model is trained while encoder model weights are frozen.


In \cite{Alvarez2019}, the encoder model is trained to predict phoneme-level labels provided from LVCSR. Both encoder and decoder models use CE-loss defined in Eq (1) and (2), where $\mathbf{X}_t = [ \mathbf{x}_{t - C_l}, \cdots , \mathbf{x}_{t}, \cdots, \mathbf{x}_{t + C_r}] $, $\mathbf{x}_{t}$ is spectral feature of d-dimension,  $y_i(\mathbf{X_t},W)$ stands for $i th$ dimension of network's softmax output, W is network weight, and ${c_t}$ is a frame-level label at frame t. 


\begin{equation}
\label{eq:ce-loss-1}
\lambda_{t}(\mathbf{W}) = - \log y_{i}(\mathbf{X}_{t}, \mathbf{W}), \:\:\;\; where \; i=c_t
\end{equation}

\begin{equation}
\label{eq:ce-loss-2}
L(W) = \sum_{t} \lambda_{t} (\mathbf{W})
\end{equation}



In \cite{Alvarez2019}, target label sequence consists of \emph{intervals} of repeated labels which we call runs. These label runs define clearly defined intervals where a model should learn to generate strong activation in label output dimension. While such model behavior can be trained end-to-end, the labels need to be provided from a LVCSR system which is typically non-end-to-end system \cite{Speech12}. The timing and accuracy of labels from LVCSR system can limit the accuracy of the trained model.


\subsection{Smoothed Max Pooling Loss}
\label{SMPloss}
Instead of \emph{interval} based CE-loss, we propose temporal max pooling operation to avoid specifying exact activation position (timing) from supervised labels. We also propose to apply temporal smoothing on the logits of frames before max pooling operation. \cite{Sun2016MaxpoolingLT} also explores max pooling loss, where one specifies a window of max pooling in the time domain, and computes CE loss only with the logit of the frame with maximum activation. However, with such simple max pooling loss, the learned activation tends to resemble a delta function, whose peak values tend to be unstable under small variation and temporal shift of audio. By introducing temporal smoothing on logits before max pooling, the model learns temporally smooth activation and stable peak values.
Eqs.(\ref{eq:mp-loss-1}) to (\ref{eq:mp-loss-5}) define the smoothed max pooling loss.

\begin{equation}
\label{eq:mp-loss-1}
Loss = Loss^+ + Loss^-	
\end{equation}

\begin{equation}
Loss^+ = \sum_{i=1}^{n}[-\log \tilde{y}_i(\mathbf{X_{m(i)}},W) ]
\label{eq:mp-loss-2}
\end{equation}

\begin{equation}
m(i)= \underset{t \in [\tau^{start}_i,\tau^{end}_i]}{argmax} \: \log \tilde{y}_i(\mathbf{X_t},W)
\label{eq:mp-loss-3}
\end{equation}

\begin{equation}
\tilde{y}_i(\mathbf{X_t},W)= s(t) \bigotimes y_i(\mathbf{X_t},W)
\label{eq:mp-loss-4}
\end{equation}

\begin{equation}
Loss^- = \sum_{t \in \tau^c}[-\log y_{c_t}(\mathbf{X_t},W)]
\label{eq:mp-loss-5}
\end{equation}

Where s(t) is a smoothing filter, $\bigotimes$ is a convolution over time and $[\tau^{start}_i, \tau^{end}_i]$ defines the interval of $i th$ max pooling window. $\tau^c$ is a set of frames \emph{not} included in any of the max pooling windows.

\subsubsection{Smoothed Max Pooling Loss for Decoder}
\label{SMPdecoder}
In our proposed approach, the decoder submodel is trained to generate strong activation on output dimension 1 near end of keyword. 
Eqs.(\ref{eq:mp-loss-1}) to (\ref{eq:mp-loss-5}) (with the number of positive targets n=1) and (\ref{eq:mp-loss-decoder-1}) to (\ref{eq:mp-loss-decoder-2}) define the loss for the decoder submodel. 

\begin{equation}
\label{eq:mp-loss-decoder-1}
\tau^{d\_start}_1 = \omega_{end} + \text{\emph{offset}}^D - win\_size^D 
\end{equation}

\begin{equation}
\label{eq:mp-loss-decoder-2}
\tau^{d\_end}_1 = \tau^{d\_start}_1 + win\_size^D 
\end{equation}

\begin{figure}
	\centering
	\includegraphics[width=\columnwidth]{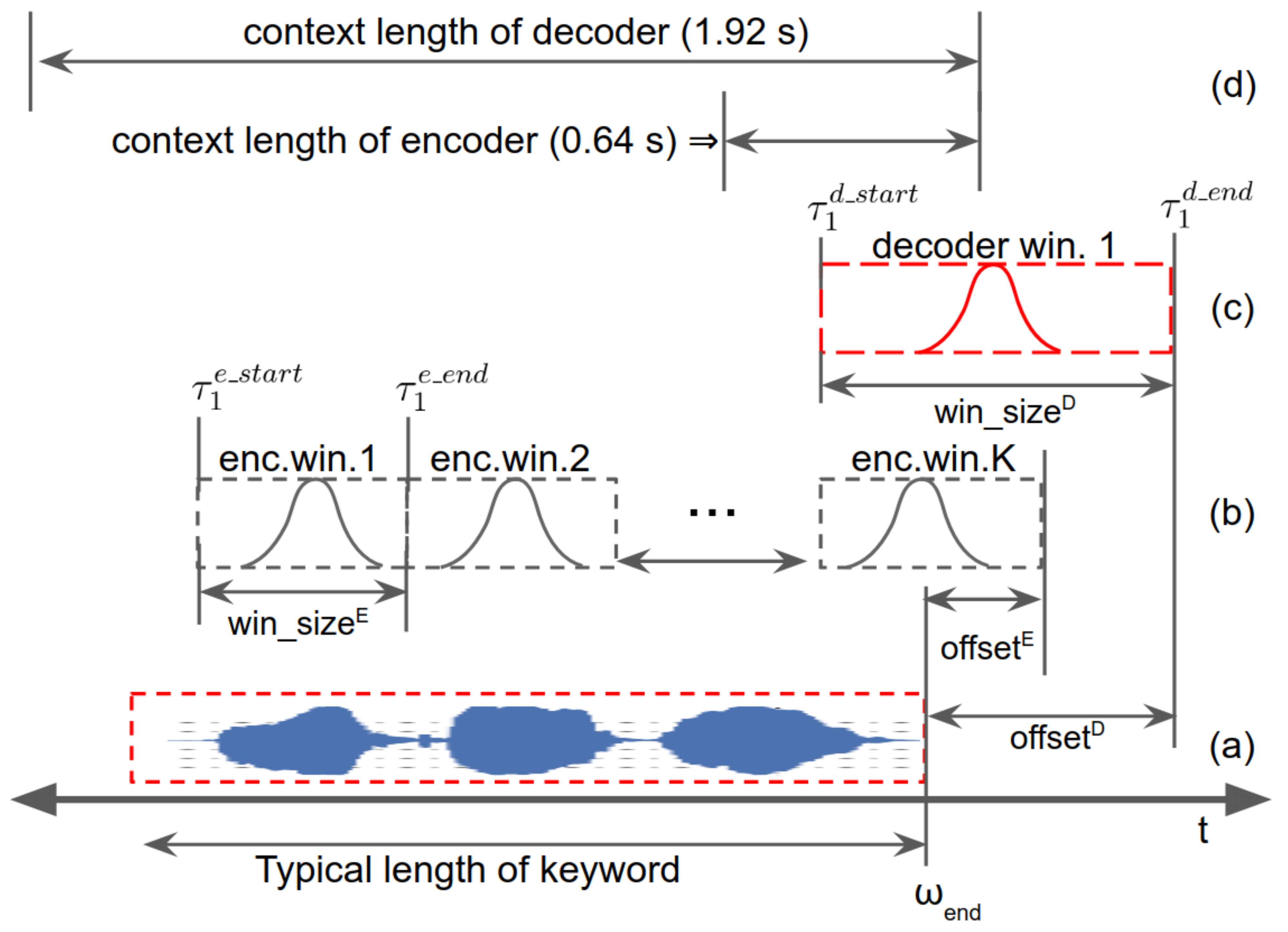}
	\caption{Relationship between max pooling windows and keyword endpoint.
(a) row shows an example of keyword audio length and endpoint Wend. 
(b) row shows encoder max pooling windows and expected activations.
(c) row shows decoder max pooling window and expected activation.
(d) row shows length of observable context for encoder and decoder.}
	\label{fig:placement}
\end{figure}

Where $\text{\emph{offset}}^D$, and $win\_size^D$ are tunable parameters. 
$\omega_{end}$ is an end-point of the expected keyword interval. Due to the nature of max pooling, the max pooling loss values are not sensitive to the exact value of $\omega_{end}$ as long as the window $[\tau^{d\_start}_1, \tau^{d\_end}_1]$  includes actual end-point of the keyword. By defining the interval long enough, the model can learn optimal position of strongest activation in a \emph{semi-supervised} manner. For current work, we used word level alignment from \cite{Speech12} to get $\omega_{end}$, but it can be computed from output of existing detection model such as \cite{Alvarez2019}. Fig. \ref{fig:placement}  (a) and (c) visualizes relationship between keyword and decoder pooling window.

\subsubsection{Smoothed Max Pooling Loss for Encoder}
\label{SMPencoder}
Unlike \cite{Sun2016MaxpoolingLT} where only decoder level output is trained with max pooling, we propose training encoder level output also using smoothed max pooling. In our method, encoder model learns a sequence of sound-parts that constitute a keyword in a semi-supervised manner. This can be done by placing K max-pooling windows sequentially over expected keyword location and define a max pooling loss at each window (Fig. \ref{fig:placement}(b)). The number of windows (n=K) and $win\_size^E$ are tuned such that K approximates number of distinguishable sound parts (i.e. phonemes), and $K*win\_size^E$ matches the average length of the keyword. Eqs.(\ref{eq:mp-loss-1}) to (\ref{eq:mp-loss-5}) and (\ref{eq:mp-loss-encoder-1}) to (\ref{eq:mp-loss-encoder-2}) define such encoder loss.

\begin{equation}
\label{eq:mp-loss-encoder-1}
\tau^{e\_start}_i = \omega_{end} + \text{\emph{offset}}^E - win\_size^E * (K-i+1), \;  \; i = 1\sim K
\end{equation}

\begin{equation}
\label{eq:mp-loss-encoder-2}
\tau^{e\_end}_i = \tau^{e\_start}_i + win\_size^E, \:\:\:\;\;\;  \; i = 1\sim K
\end{equation}

Where $\text{\emph{offset}}^E$ and $win\_size^E$ are also tunable parameters. 
Fig.\ref{fig:placement} (a) and \ref{fig:placement} (b) show the relationship between expected keyword and pooling windows. Both the encoder and the decoder models are trained jointly using loss in (\ref{eq:loss-eqn-combined}). The tunable parameter $\alpha$ controls the relative importance of each loss.


\begin{equation}
\label{eq:loss-eqn-combined}
 Total\_loss = \alpha * Loss^E + Loss^D
\end{equation}

\section{Experimental setup}
\label{sec:setup}

We compare the model trained with the new smoothed max pooling loss on encoder/decoder architecture with the baseline in \cite{Alvarez2019}. Both the baseline and the proposed model have the same architecture. Only the training losses are different. Details of the setup are discussed below. 


\subsection{Front-end}
\label{frontend}
We used the same frontend feature extract as the baseline \cite{Alvarez2019} in our experiments. The front-end extracts and stacks a 40-d feature vector of log-mel filter-bank energies at each frame and stacks them to generate input feature vector $\mathbf{X}_t$. Refer to \cite{Alvarez2019} for further details.


\subsection{Model setup}
\label{setup}
We selected E2E\_318K architecture in \cite{Alvarez2019} as the baseline and use the same structure for testing all other models. As shown in Fig. \ref{fig:topology}, the model has 7 SVDF layers and 3 linear bottleneck dense layers. For detailed architectural parameters, please refer to \cite{Alvarez2019}. We call the baseline model as Baseline\_CE\_CE where encoder and decoder submodels are trained with CE loss. We call the proposed model as Max4\_SMP\_SMP where both encoder and decoder submodels are trained by SMP (smoothed max pooling) loss.

We also performed ablation study by testing other models that use different losses. Table \ref{tab:models} summarizes all the tested models. Model Max1--Max3 uses SMP (smoothed max pooling) loss for the decoder, but uses different losses for the encoder.  Max1\_CTC\_SMP used CTC loss to train the encoder. Standard CTC loss function from Tensorflow \cite{Graves2006ConnectionistTC} was used. CTC loss doesn’t need alignments, but it learns \emph{peaky} activations whose peak values are not highly stable. Max2\_NA\_SMP has no encoder loss (i.e. $\alpha=0$), s.t. the entire network is trained by decoder loss only. Max3\_CE\_SMP used baseline CE loss for encoder. Model Max4--Max7 are tested to measure the importance of the \emph{smoothing} operation. MP means max pooling without smoothing (i.e. $s(t)=1$).


\begin{table}[h!]
	\vspace{-4mm}
	\begin{center}
		\caption{Summary of various models tested}
		\label{tab:models}
		\begin{tabular}{l|r|r}
			\textbf{Models} & encoder loss & decoder loss\\
			\hline
			Baseline\_CE\_CE & cross entropy  & cross entropy\\
			Max1\_CTC\_SMP   & CTC loss & smoothed MP \\
			Max2\_NA\_SMP    & -- & smoothed MP \\
			Max3\_CE\_SMP    & cross entropy & smoothed MP \\
			\hline
			Max4\_SMP\_SMP   & smoothed MP & smoothed MP \\
			\hline
			Max5\_MP\_SMP    & MP & smoothed MP \\
			Max6\_SMP\_MP    & smoothed MP & MP \\
			Max7\_MP\_MP     & MP & MP\\
		\end{tabular}
	\end{center}
	\vspace{-4mm}
\end{table}


For the decoder SMP(smoothed max pooling) loss, we used truncated Gaussian as the smoothing filter $s(t)$ with $\mu=0$, $\sigma=9$ frames (90ms) and truncated length 21 frames. Max pooling window of size 60 frames (600ms) with $\text{\emph{offset}}^D = 40$ frames (400ms) is used. For the encoder SMP loss, we used truncated gaussian with $\mu=0$, $\sigma=4$ frames and truncated length 9. Encoder max pooling windows have size of 20 frames with $\text{\emph{offset}}^E = 40$ frames. These windows are placed sequentially in 40 frames interval.



\subsection{Dataset}
\label{ssec:data}
The training data consists of 2.1 million anonymized utterances with the keywords “Ok Google” and “Hey Google”. Data augmentation similar to \cite{Alvarez2019} has been used for better robustness.

Evaluation is done on four data sets separate from training data, representing diverse environmental conditions -- \emph{Clean non-accented} set contains 170K non-accented English utterances of keywords in quiet condition. \emph{Clean accented} has 138K English utterances of keyword with Australian, British, and Indian accents in quiet conditions. 
\emph{Query logs} contains 58K utterances from anonymized voice search queries.
\emph{In-vehicle} set has 144K utterances with the keywords recorded inside cars while driving, which includes significant amount of noises from road, engine, and fans. All sets are augmented with 64K negative utterances, which are re-recorded TV noise.

\section{Results}
\label{sec:results}

To show effectiveness of the proposed approach, we evaluated false-reject (FR) and false-accept (FA) tradeoff across various models described in Section \ref{sec:setup}. All models are converted to inference models using TensorFlow Lite’s quantization \cite{TfliteSvdfHybridQuant}.

\begin{figure}[htb]
	\begin{minipage}[b]{\linewidth}
		\centering
		\centerline{\includegraphics[height=137pt]{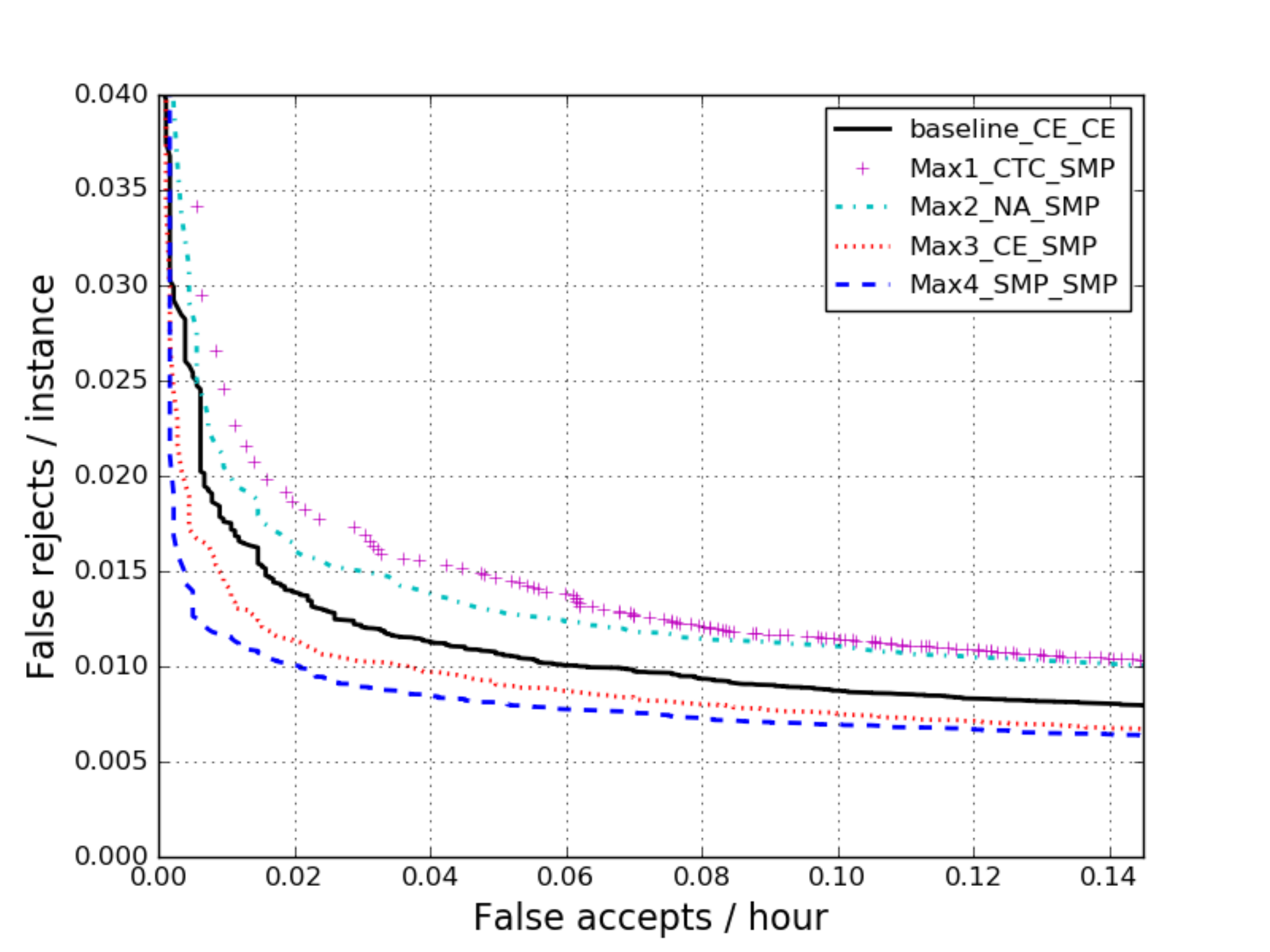}}
		\centerline{(a) Clean non-accented}\medskip
	\end{minipage}
	\hfill
	\begin{minipage}[b]{\linewidth}
		\centering
		\centerline{\includegraphics[height=137pt]{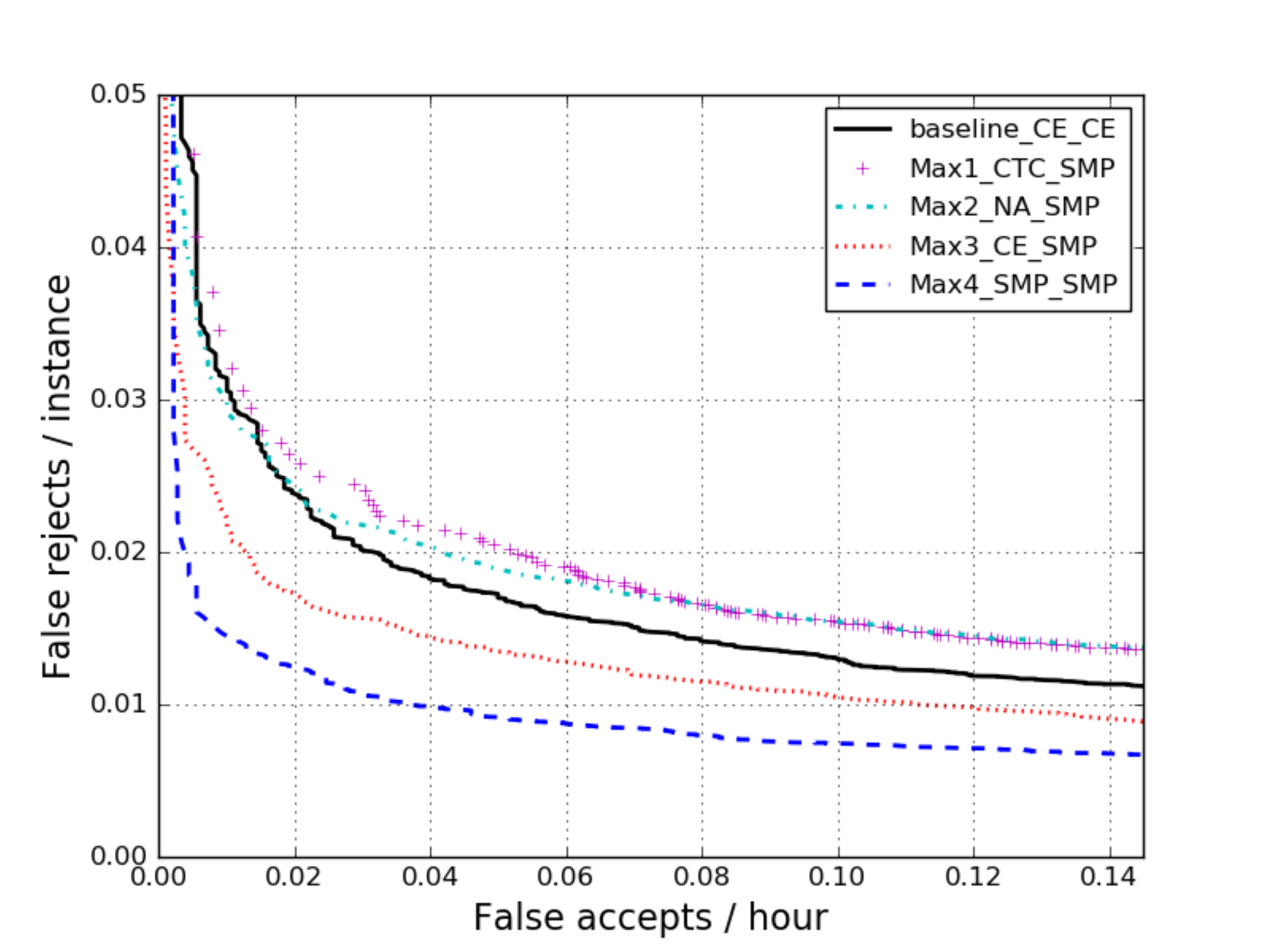}}
		\centerline{(b) Clean accented}\medskip
	\end{minipage}
	\begin{minipage}[b]{\linewidth}
		\centering
		\centerline{\includegraphics[height=137pt]{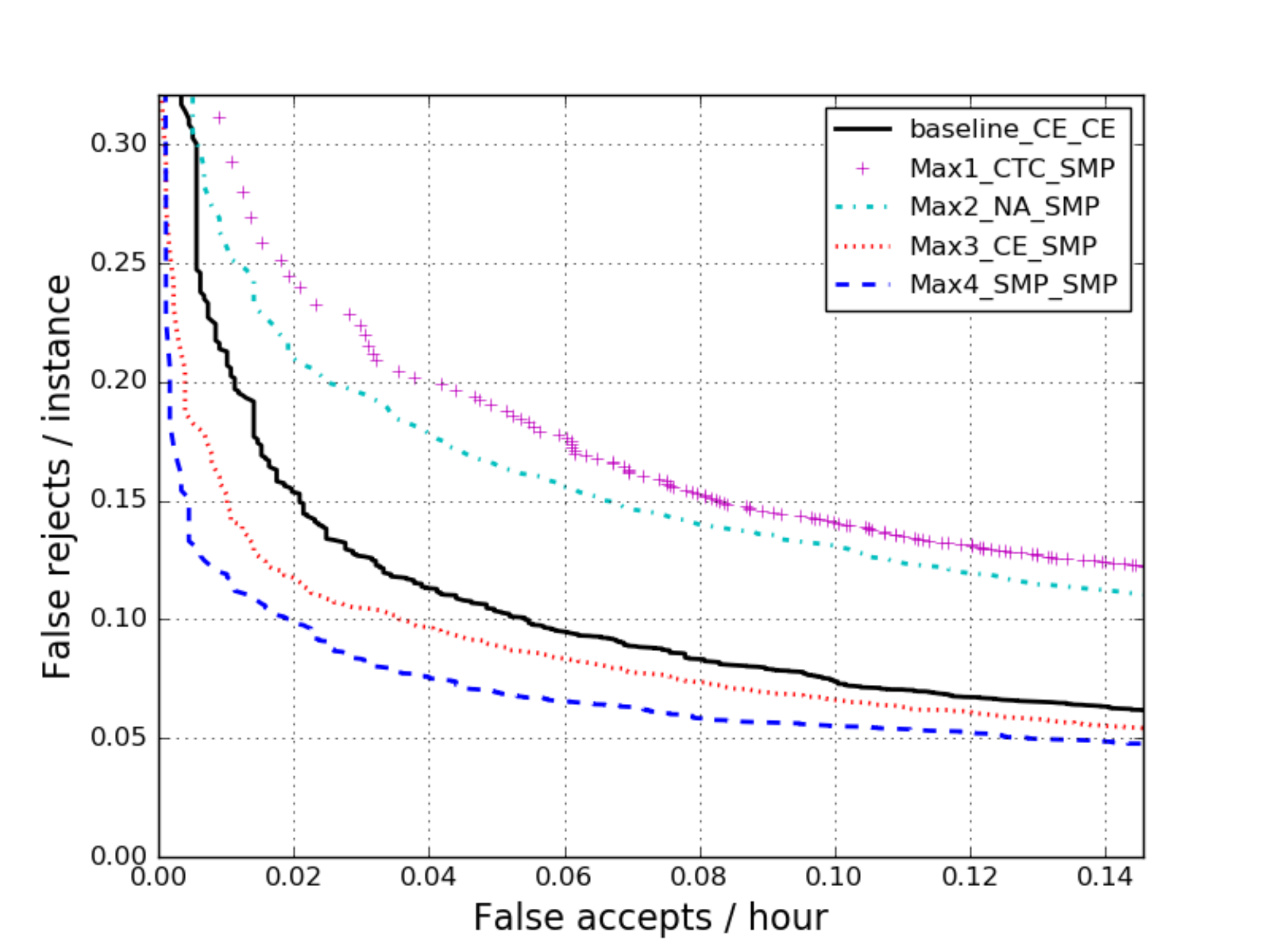}}
		\centerline{(c) Anonymous query logs}\medskip
	\end{minipage}
	\hfill
	\begin{minipage}[b]{\linewidth}
		\centering
		\centerline{\includegraphics[height=137pt]{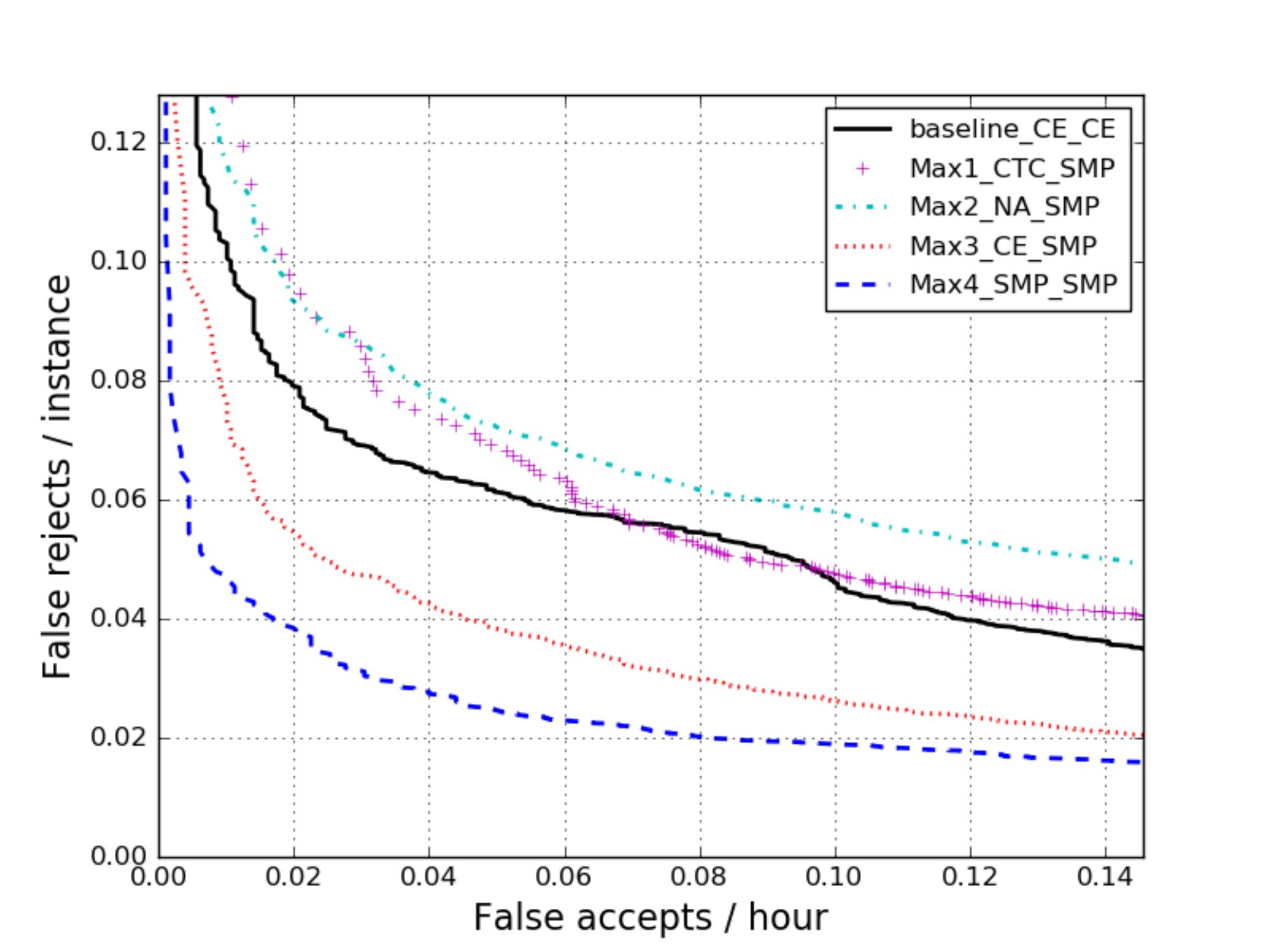}}
		\centerline{(d) Noisy inside vehicle}\medskip
	\end{minipage}
	\caption{ROC curves of models with various loss types and conditions}
	\label{fig:rocA}
\end{figure}

\begin{figure}[htb]
	\begin{minipage}[b]{\linewidth}
		\centering
		\centerline{\includegraphics[height=137pt]{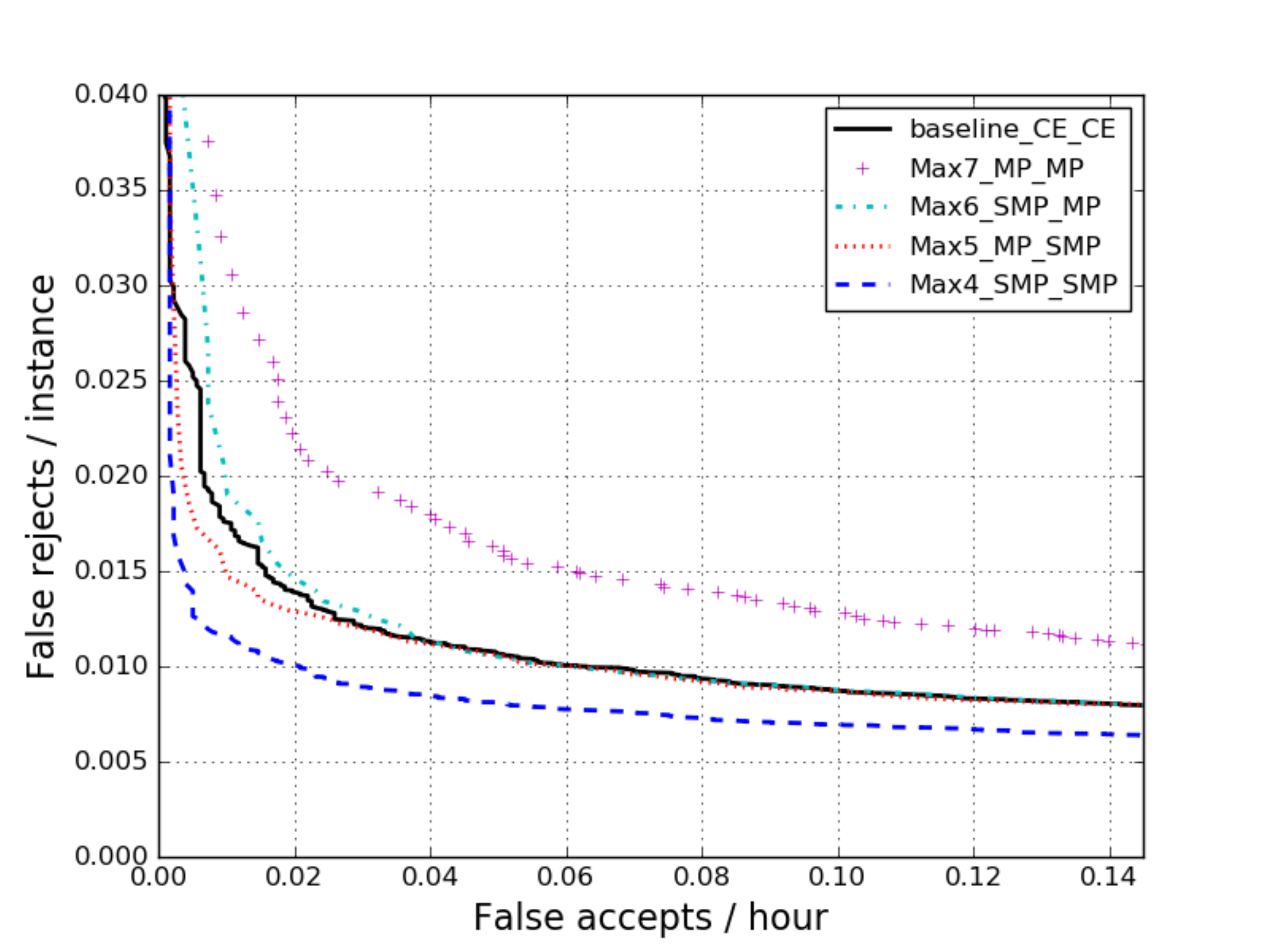}}
		\centerline{(a) Clean non-accented}\medskip
	\end{minipage}
	\hfill
	\begin{minipage}[b]{\linewidth}
		\centering
		\centerline{\includegraphics[height=137pt]{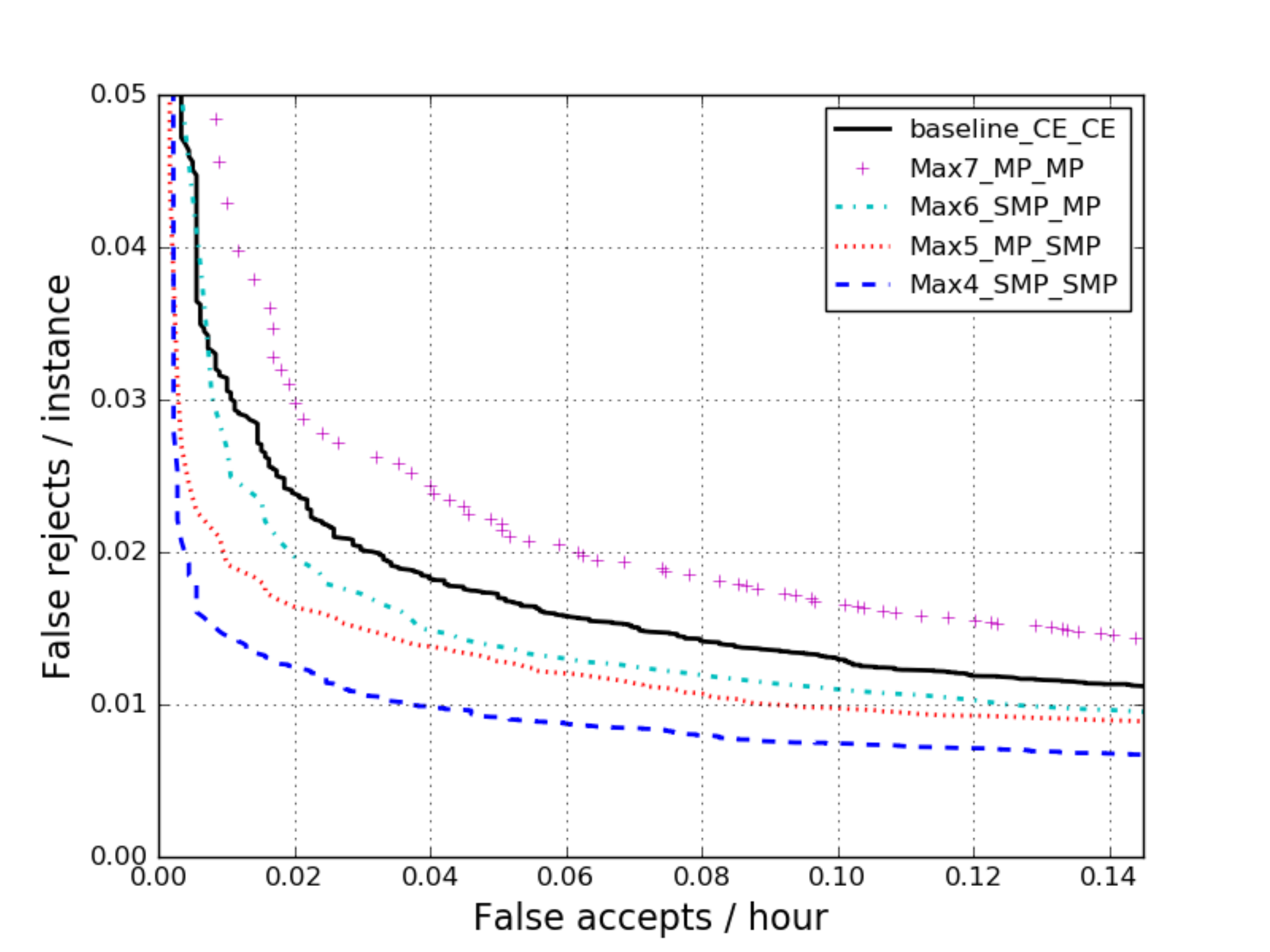}}
		\centerline{(b) Clean accented}\medskip
	\end{minipage}
	\begin{minipage}[b]{\linewidth}
		\centering
		\centerline{\includegraphics[height=137pt]{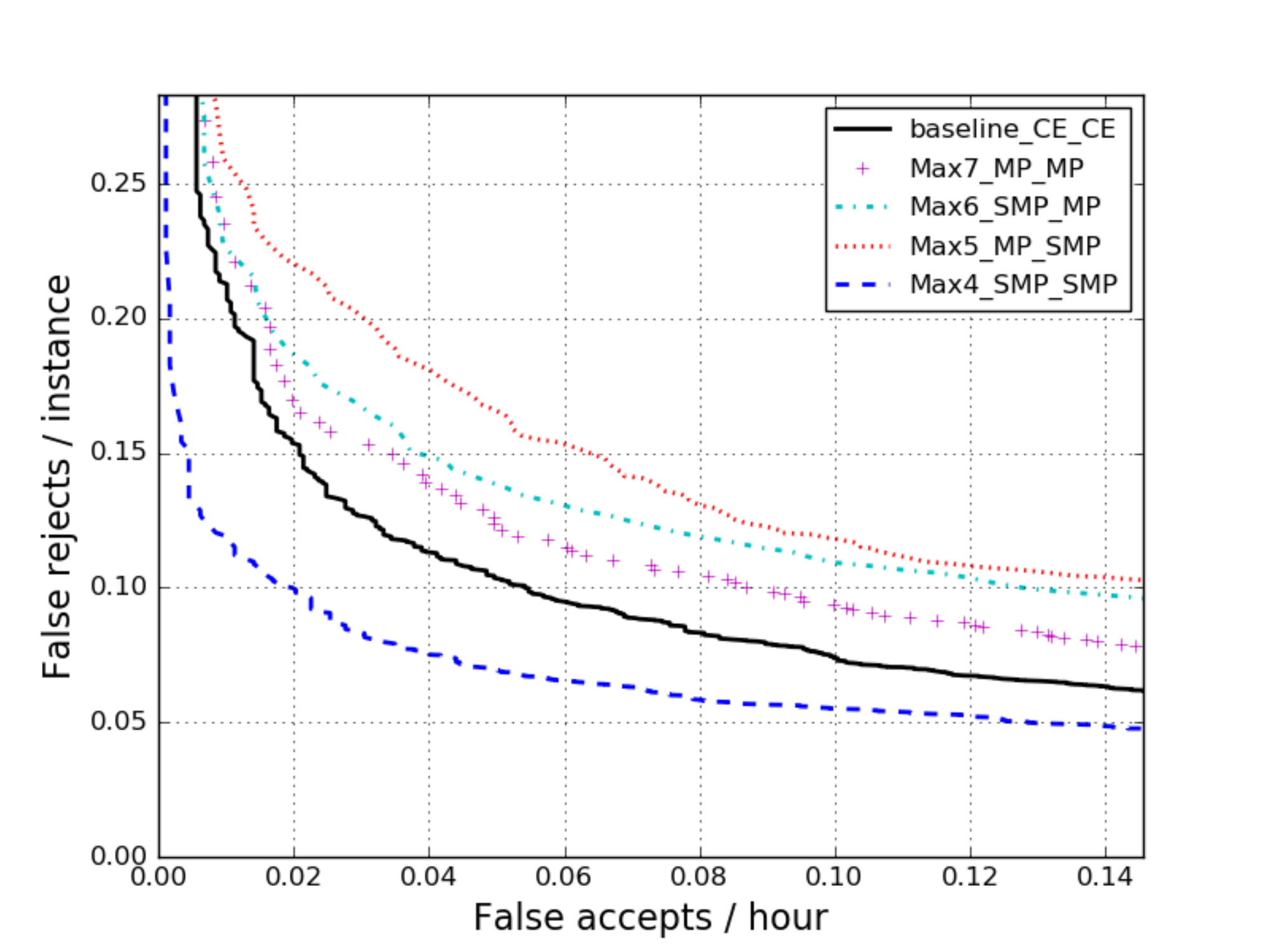}}
		\centerline{(c) Anonymous query logs}\medskip
	\end{minipage}
	\hfill
	\begin{minipage}[b]{\linewidth}
		\centering
		\centerline{\includegraphics[height=137pt]{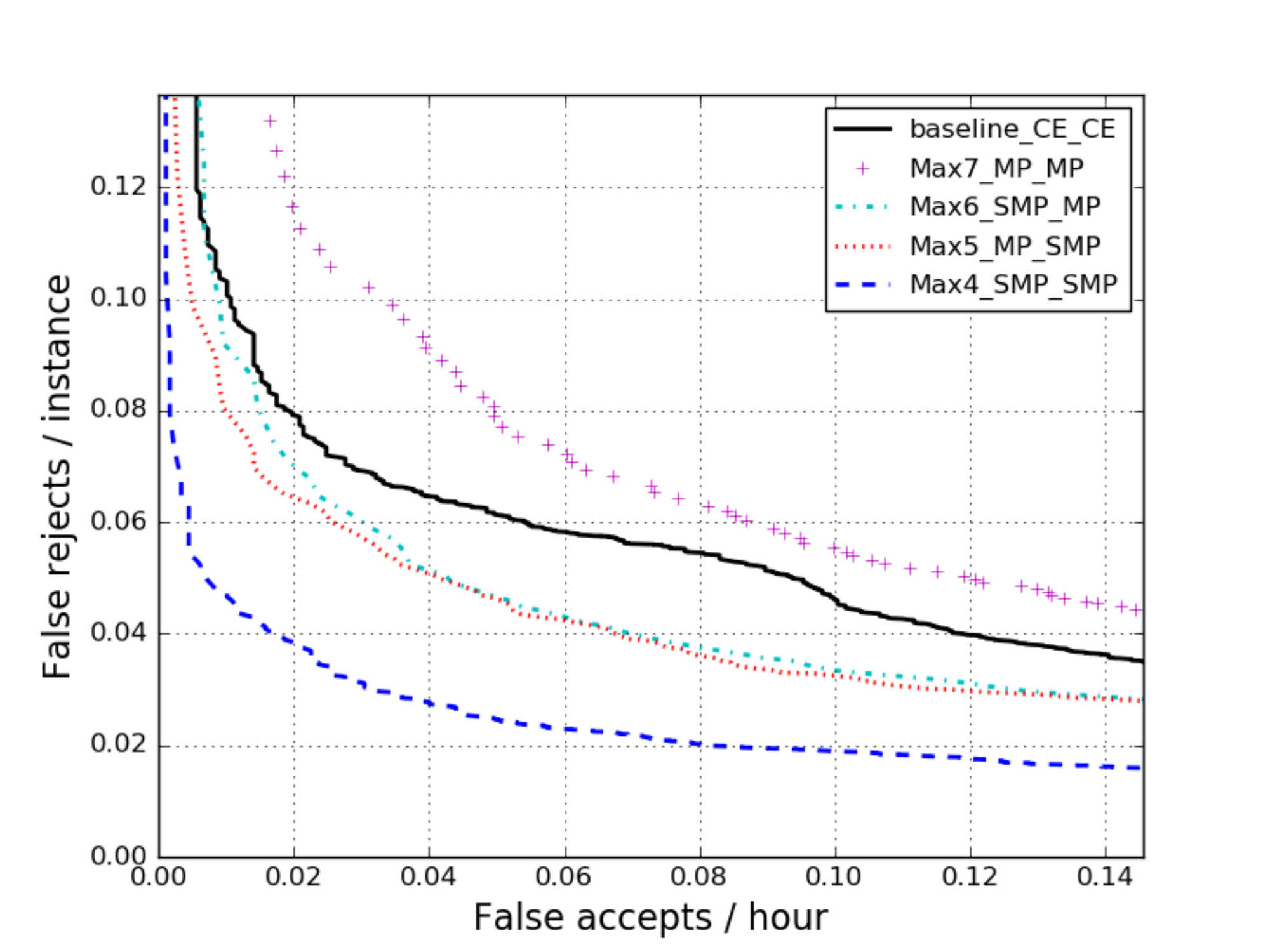}}
		\centerline{(d) Noisy inside vehicle}\medskip
	\end{minipage}
	\caption{ROC curves of models with various smoothing options}
	\label{fig:rocB}
\end{figure}

Table \ref{tab:perfdiffA} summarizes FR rates of models in Fig.\ref{fig:rocA} and \ref{fig:rocB} at selected FA rate (0.1 FA per hour measured on 64K re-recorded TV noise set). Fig.\ref{fig:rocA} shows the ROC curves of various models (baseline, Max1--Max4) across different conditions. Figure \ref{fig:rocB} shows the ROC curves of Max4--Max7 models across different conditions.

Across model types and evaluation conditions {Max4\_SMP\_SMP} shows the best accuracy and ROC curve. {Max3\_CE\_MP} model also performs better than the baseline but not as good as Max4. Other variations {Max2} (has only decoder loss) and {Max1} (has CTC encoder loss) performed worse than baseline.
Comparison among models with max pooling and different smoothing options (Fig.\ref{fig:rocB}) shows that {Max4\_SMP\_SMP} (smoothed max poling on both encoder and decoder) performs the best and outperforms {Max7}(no smoothing on encoder and decoder max pooling loss). Especially the proposed Max4 model reduces FR rate to \emph{nearly half} of the baseline in \emph{clean accented} and \emph{noisy inside-vehicle} conditions, where it's more difficult to obtain training data with accurate alignments.

\begin{table}[h!]
	\vspace{-4mm}
	\begin{center}
		\caption{FR rate of models with various loss types at 0.1 FA/h}
		\label{tab:perfdiffA}
		\begin{tabular}{l|r|r|r|r}
			\textbf{Models} & Non Acc. & Accented & Q. Logs & Vehicle \\
			\hline
			Baseline\_CE\_CE & 0.97\% & 1.48\% & 8.89\% & 5.62\%  \\
			Max1\_CTC\_SMP   & 1.26\% & 1.73\% & 16.1\% & 5.57\%  \\
			Max2\_NA\_SMP    & 1.18\% & 1.68\% & 14.5\% & 6.41\% \\
			Max3\_CE\_SMP    & 0.83\% & 1.20\% & 7.79\% & 3.21\%  \\
			\hline
			Max4\_SMP\_SMP   & 0.74\% & 0.82\% & 6.13\% & 2.58\%  \\
			\hline
			Max5\_MP\_SMP   & 0.96\% & 1.12\% & 14.2\% & 3.91\%  \\
			Max6\_SMP\_MP    & 0.96\% & 1.24\% & 12.5\% & 4.01\% \\
			Max7\_MP\_MP    & 1.46\% & 1.93\% & 11.0\% & 6.83\%  \\		
		\end{tabular}
	\end{center}
	\vspace{-8mm}
\end{table}

\section{Conclusion}
\label{sec:conclusion}
We presented smoothed max pooling loss for training keyword spotting model with improved optimizability. Experiments show that the proposed approach outperforms the baseline model with CE loss by relative 22\%--54\% across a variety of conditions. Further, we show that applying smoothing before max pooling is highly important for achieving accuracy better than the baseline. The proposed approach provides further benefits of reducing dependence on LVCSR to provide phoneme level alignments, which is desirable for embedded learning scenarios, like on-device learning \cite{Fed17}\cite{Leroy2018FederatedLF}.

\clearpage
\bibliographystyle{IEEEbib}
\bibliography{refs}

\end{document}